\documentclass[runningheads]{llncs}
\usepackage{graphicx}
\usepackage{todonotes}
\usepackage{amsmath}
\usepackage{siunitx}
\usepackage{multirow}
\usepackage{algorithm}
\usepackage{subcaption}
\usepackage{adjustbox}
\usepackage{gensymb}

\usepackage{algpseudocode}
\begin{document}
\title{Multitemporal Aerial Image Registration Using Semantic Features\thanks{A. Gupta is funded by a Scholarship from the Department of Electrical and Electronic Engineering, The University of Manchester and the ACM SIGHPC/Intel Computational and Data Science Fellowship }}
\titlerunning{Multitemporal Aerial Image Registration Using Semantic Features}
%
\author{Ananya Gupta\orcidID{0000-0001-6743-1479} \and
Yao Peng\orcidID{0000-0002-3501-2146}\and
Simon Watson\orcidID{0000-0001-9783-0147} \and
Hujun Yin\orcidID{0000-0002-9198-5401}}
\authorrunning{A. Gupta et al.}
%
\institute{The University of Manchester, Manchester, United Kingdom \\
\email{\{ananya.gupta, yao.peng, simon.watson, hujun.yin\}@manchester.ac.uk}}
\maketitle              
\begin{abstract}
A semantic feature extraction method for multitemporal high resolution aerial image registration is proposed in this paper. These features encode properties or information about temporally invariant objects such as roads and help deal with issues such as changing foliage in image registration, which classical handcrafted features are unable to address. These features are extracted from a semantic segmentation network and have shown good robustness and accuracy in registering aerial images across years and seasons in the experiments. 

\keywords{Image Registration  \and Semantic Features \and Convolutional Neural Networks}
\end{abstract}
\section{Introduction}
Image registration is widely used for aligning images of the same scene. These images can be taken at different times, by different sensors and viewpoints and hence have appearance differences due to the varying imaging conditions~\cite{Zitova2003}. Registration is particularly useful in remote sensing for aligning multitemporal and/or multispectral imagery for tasks such as multi-sensor data fusion and change detection~\cite{XiaolongDai1998}. It can also be used for Unmanned Aerial Vehicle (UAV) localisation by matching online UAV (query) images with the corresponding aerial (reference) images~\cite{Costea2016}. Image registration is also used in medical diagnosis for tumour monitoring or analysis of treatment effectiveness~\cite{Zitova2003}.

Image registration methods can be broadly divided into two categories: area-based and feature-based~\cite{Zitova2003}. Area-based methods try to match corresponding patches in two images using similarity based metrics~\cite{Zagoruyko}. Feature-based methods, on the other hand, extract salient features from two images and then try to find corresponding features to estimate the transformation between the two images. Area-based methods are typically limited to images with different translations and/or minor rotations but cannot deal with scale variations. They also do not perform any structural analysis, hence multiple smooth areas can lead to incorrect correlation. Feature-based methods match more distinctive locations and are more robust. Hence, feature-based algorithms are more commonly used in registration, especially in remote sensing for matching areas with distinctive objects such as buildings and roads.

A number of popular feature-based methods are based on variants of the scale-invariant feature transform (SIFT) descriptors~\cite{Lowe2004,Yang2017b,Sedaghat2011}. For instance, SAR-SIFT was developed to register synthetic aperture radar (SAR) images~\cite{Dellinger2015}. Fast Sample Consensus with SIFT was proposed as an improvement to random sample consensus to improve the number of correct correspondences for image registration~\cite{YueWu2015}. A coarse to fine registration strategy based on SIFT and mutual information was proposed to achieve good outlier removal~\cite{MaoguoGong2013}. SIFT provides robust features in terms of translation and scale invariance. However, they can only take local appearance into account and lose global consistency.

Recently, deep learning is also being used to extract features from aerial imagery~\cite{Zhang2016a}. A method for combining SIFT features with Convolutional Neural Network (CNN) features has been developed for remote sensing image registration~\cite{Ye2018}. CNN features have also been used for registration of multitemporal images~\cite{Yang2018}. However, these methods were trained for image classification and do not encode fine-grained information about objects, hence they do not work well in registering high resolution images with fine details.

In this paper, we focus on registering multitemporal high resolution nadir aerial images that have large variations due to changing seasons, lighting conditions, etc. We propose to use semantic features extracted from a segmentation network for the purpose of aerial image registration. These segmentation-based semantic features (SegSF), as compared to handcrafted and classification-based CNN features, are more finely localised and more discriminative for multitemporal registration. 

\section{Methodology}

The proposed methodology comprises of two main steps, SegSF extraction and class-specific feature matching, detailed as follows.

\subsection{Segmentation-Based Semantic Feature Extraction}

\begin{figure}
    \centering
    \includegraphics[width=\textwidth]{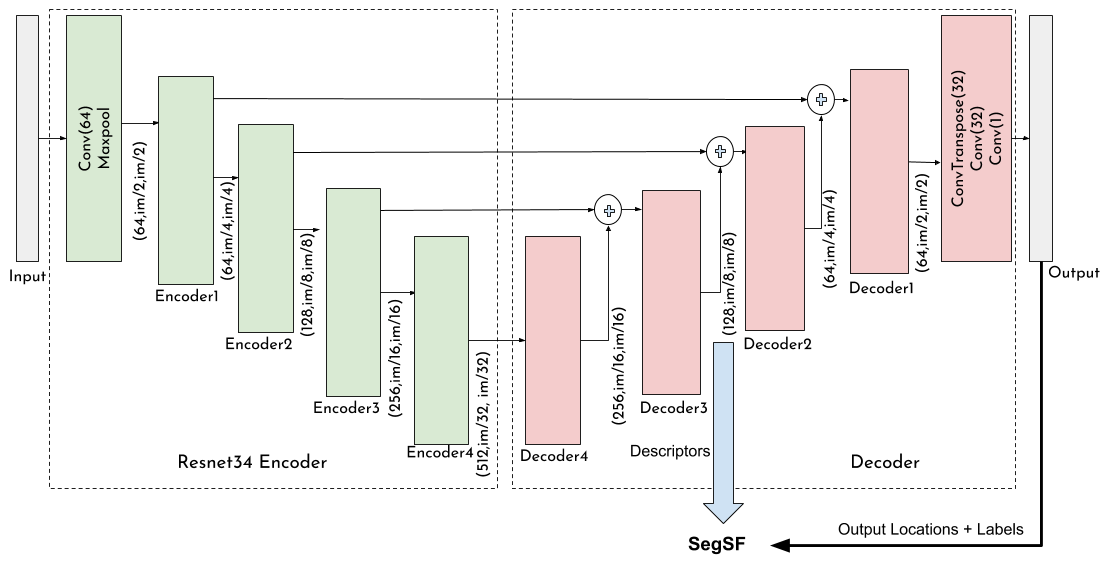}
    \caption{LinkNet34 architecture for SegSF extraction. Each block in the encoder downsamples the image by a factor of 2 and each corresponding decoder block upsamples the image as shown by the block output sizes in the figure. The image width and height are each denoted by \(im\).}
    \label{fig:semf}
\end{figure}{}

A LinkNet34~\cite{Chaurasia2018} network trained for road segmentation with aerial images as the input and binary road masks as the output is used for SegSF extraction. The network structure is shown in Fig. \ref{fig:semf}. The network is trained with a pixel-wise loss function defined by the binary cross entropy between predicted value and ground truth road mask. It provides a probability mask as output, which is converted to a binary road mask with a threshold of 0.5, so any pixels with a probability greater than 0.5 are regarded as road pixels and vice versa.

The features from the output of the `Decoder3' block shown in Fig. \ref{fig:semf} are extracted as descriptors. Additionally, keypoint locations on the image are given by the centers of the effective receptive field of the descriptors and can be calculated using Equations 1-4.

\begin{gather}
    j_{l} = j_{l-1} * s_l \label{eq:jump} \\ 
    rf_l = rf_{l-1} + \left(k_l-1\right) * j_{l-1} \label{eq:rf} \\
    start_l = start_{l-1} + \Big( \frac{k_l-1}{2} - p_l \Big) * j_{l-1} \\
    outputLoc_l = featureLoc_l*rf_l + start_l
\end{gather}
where subscripts \(l\) and \(l-1\) are layer indices, \(s\) is the stride, \(p\) gives the padding size, \(k\) is the convolution kernel size and \(j\) is the "jump" or the effective stride for each layer as compared to the input. So the jump for the first layer is the same as its stride. The effective receptive field has size of \(rf\) and \(start\) is the center coordinate of the receptive field of the first feature. Corresponding output locations (\(outputLoc\)) and hence labels can then be assigned to feature descriptors using Eq. 4, where \(featureLoc_l\) indexes over all features in the feature map output from layer \(l\).

In practice, since the encoder is a ResNet34~\cite{He2015}, the values of  padding, kernel and stride are chosen so that the effective jump only changes between two encoder blocks and the \(start\) value is always 0.5.

SegSF features are defined by three components: class label, descriptor and  keypoint location. Each descriptor-keypoint pair is assigned a class label based on the location of the keypoint on the segmentation output, hence adding semantic knowledge to the feature descriptor.

\subsection{Feature Matching}

All SegSF descriptors are L2-normalised individually and their dimensionality is reduced to 100 using class-specific PCA, further followed by L2 normalisation to obtain the final descriptor. 

Matching class descriptors between the query and reference images are found using nearest neighbour search in the per-class descriptor space where Euclidean distance is used as the distance metric. The correctness of the extracted correspondences is estimated using Lowe's ratio test~\cite{Lowe2004}, where the match is assumed to be correct if the distance ratio between the first neighbour and the second neighbour is less than 0.7.

The keypoints of the feature matches are then used to estimate the homography matrix between the two images using random sample consensus~\cite{Fischler1981}. This feature matching process has been given as Algorithm \ref{alg:semf}.

\begin{algorithm}
\caption{SegSF matching}\label{alg:semf}
\hspace*{\algorithmicindent} \textbf{Inputs}: Query Image, Reference Image \textit{query\_image, ref\_image} \\
\hspace*{\algorithmicindent} \textbf{Output}: Transformation Model \textit{Transform\_Model} 

\begin{algorithmic}[1]
    \State $all\_ref\_kp$ = []
    \State $all\_query\_kp$ = []
    \For{$class$ in $num\_classes$}
        \State $query\_class\_kp$, $query\_class\_des$ = GetClassFeatures($class$, $query\_image$)
        \State $ref\_class\_kp$, $ref\_class\_des$ = GetClassFeatures($class$, $ref\_image$)
        \State $query\_class\_des$.NormalisePCANormalise
        \State $ref\_class\_des$.NormalisePCANormalise
        \State $inlier\_query\_kp$, $inlier\_ref\_kp$ = KNN($query\_des$, $ref\_des$)
        \State $all\_ref\_kp$.append($inlier\_ref\_kp$)
        \State $all\_query\_kp$.append($inlier\_query\_kp$)
    \EndFor
    
    \State $Transform\_Model$ = Affine TransformRansac($all\_ref\_kp$, $all\_query\_kp$)

\end{algorithmic}
\end{algorithm}

\section{Experimental setup}

\subsection{Datasets}
The segmentation network was trained on images from different seasons and years to learn temporally invariant features. Aerial imagery datasets provided by the Australian Capital Territory Government~\cite{actmapi} for the years 2015-2018 were used for this purpose. The images in this dataset are georeferenced, orthorectified and have a ground sampling distance of \SI{10}{\centi\metre} with an expected error less than \SI{20}{\centi\metre}. An area of \SI{27.5}{\kilo\metre\squared} around Canberra was extracted for the experiments, with 90\% of the images from 2015, 2016 and 2017 being used for training and validation of the segmentation network. The remaining 10\% images from 2017 and all images from 2018 were set aside for testing.

The annotations for training were extracted from the OpenStreetMap (OSM)~\cite{Contributors2017} where all polylines marked as one of motorways, primary, residential, secondary, service, tertiary and trunk and their respective links were assumed to be roads. The OSM roads were provided in vector format and were rasterised with a width of \SI{2}{\metre} to obtain data suitable for training the segmentation network.


\subsection{Training Details}

The segmentation network was trained on image crops of \(416\times416\) from the training dataset of 2600 images. It was trained for 200 epochs with the images being augmented by random horizontal and vertical flipping with a probability of 0.5. All image pixel values were normalised between 0 and 1. The Adam optimiser~\cite{Kingma2015} with a learning rate of 0.0001 was used for optimisation. Pytorch~\cite{pytorch} was used for creating and training the neural networks.

\subsection{Testing Scheme and Metrics}

The test images from 2017 were rotated around their center point by angles of 1\degree, 2\degree, 3\degree, 4\degree, 5\degree, 10\degree, 15\degree, 20\degree, 30\degree and 40\degree. Corresponding images from the 2018 dataset were extracted and the methods were tested on their accuracy for registering these multitemporal images. Samples of the test pairs can be seen in Fig. \ref{fig:cb}. Note that the network was not trained on any images from the 2018 dataset or on any images from the testing region. 

Since the image transformation parameters were known, a per-pixel metric has been reported for the experiments. The root mean squared error (RMSE) between the pixel positions using the predicted transformation and the actual transformation was used as the error metric. The mean values of the errors over all the images for the different transformations are reported in Section \ref{sec:results}.

\section{Results}
\label{sec:results}

We compared our results with that of CNN-Reg~\cite{Yang2018} which is based on extracting multi-scale CNN features for multitemporal remote sensing image registration. They utilise features from a VGG-16~\cite{Simonyan2015} network trained on the ImageNet dataset~\cite{Russakovsky2015} for classification. We have also reported the t-values and p-values from Welch's t-test~\cite{WELCH1947} to obtain the significance of the results.

\begin{table}[]
\centering
\caption{Registration RMSE of CNN-Reg and SegSF on multi-temporal aerial imagery dataset over different rotation parameters}
\label{tab:reg_results}
\begin{adjustbox}{width=\textwidth}
\begin{tabular}{|l|l|l|l|l|l|l|l|l|l|l|}
\hline
\textbf{Method}          & \textbf{1\degree} & \textbf{2\degree} & \textbf{3\degree} & \textbf{4\degree} & \textbf{5\degree} & \textbf{10\degree} & \textbf{15\degree} & \textbf{20\degree} & \textbf{30\degree} & \textbf{40\degree} \\ \hline
SegSF& 37.03     & 37.33      & 41.93     & 46.43      & 61.51     & 70.08       & 90.63       & 131.91      & 184.06      & 372.39      \\ \hline
CNN-Reg & 88.05      & 143.51     & 213.18     & 242.806    & 287.13    & 425.63     & 522.5       & 596.35    & 715.13      & 777.53     \\ \hline
t-value & 5.11 & 8.38 & 10.87 & 11.62 & 12.41 & 17.82 & 22.92 & 20.79 & 24.21 & 10.37 \\ \hline
p-value & 1.53e-6 & 3.07e-12 & 3.81e-18 & 5.55e-19 & 9.59e-22 & 8.58e-28 & 5.68e-37 & 8.53 e-38 & 6.09e-38 & 1.46e-14 \\
\hline
\end{tabular}
\end{adjustbox}
\end{table}

\begin{figure}[]
\begin{tabular}{|c|c|c|c|}
\hline
\textbf{Reference Image} & \textbf{Query Image} & \textbf{CNN-Reg} & \textbf{SegSF}\\ & & &\\

\begin{subfigure}[b]{0.24\textwidth}
\includegraphics[width = \textwidth]{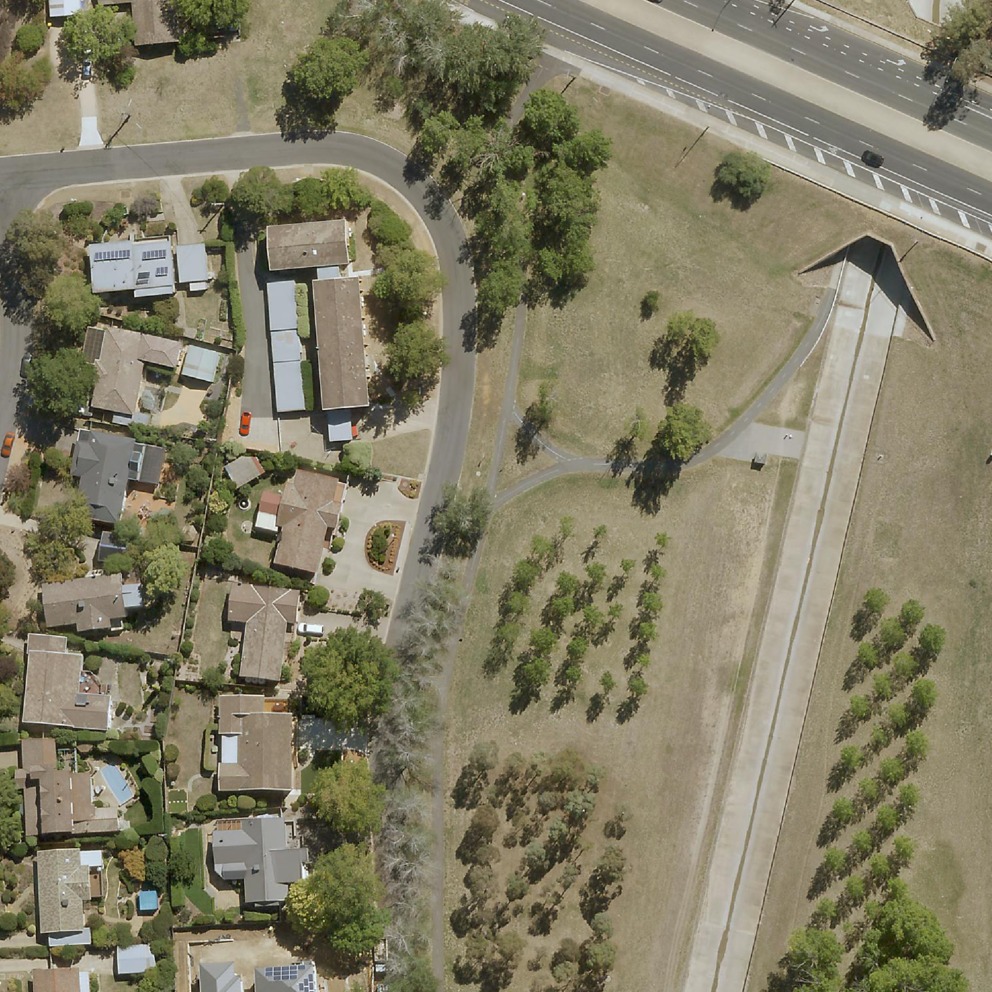} 
\end{subfigure} &
\begin{subfigure}[b]{0.24\textwidth}
\includegraphics[width = \textwidth]{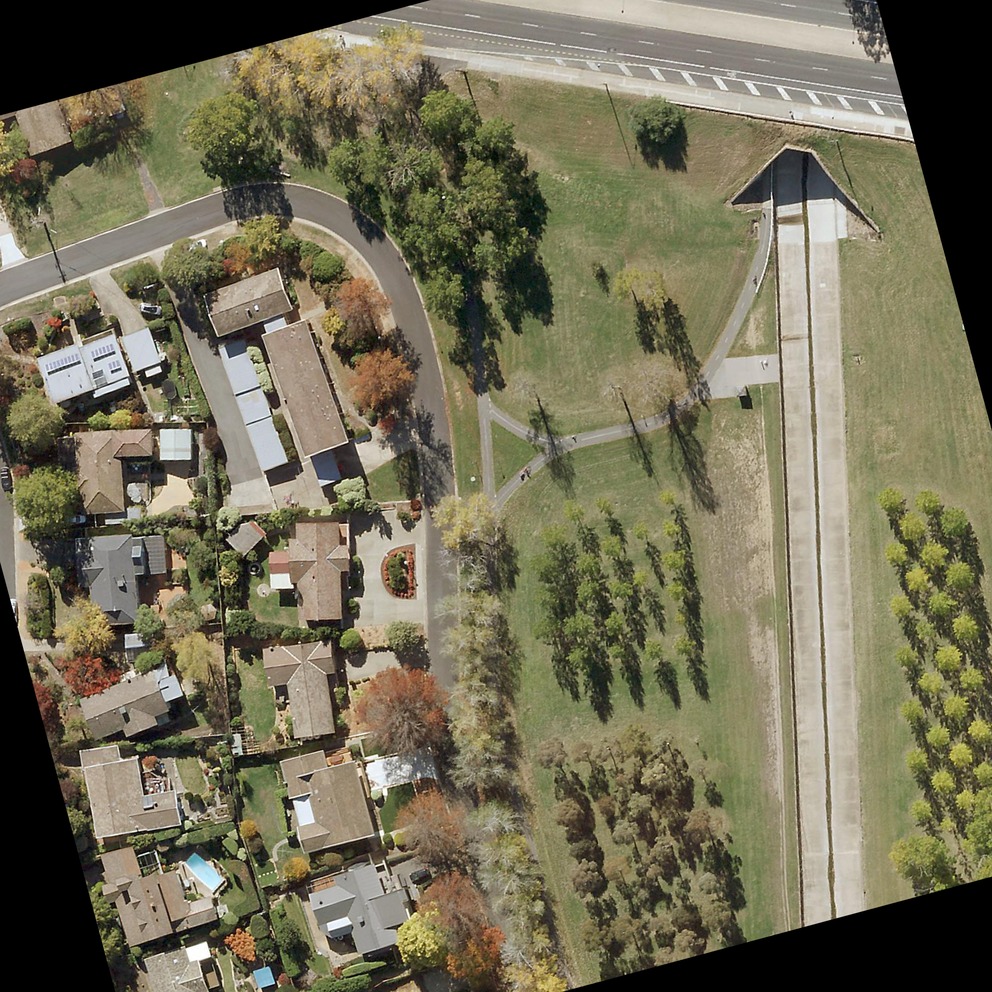} 
\end{subfigure} &
\begin{subfigure}[b]{0.24\textwidth}
\includegraphics[width = \textwidth]{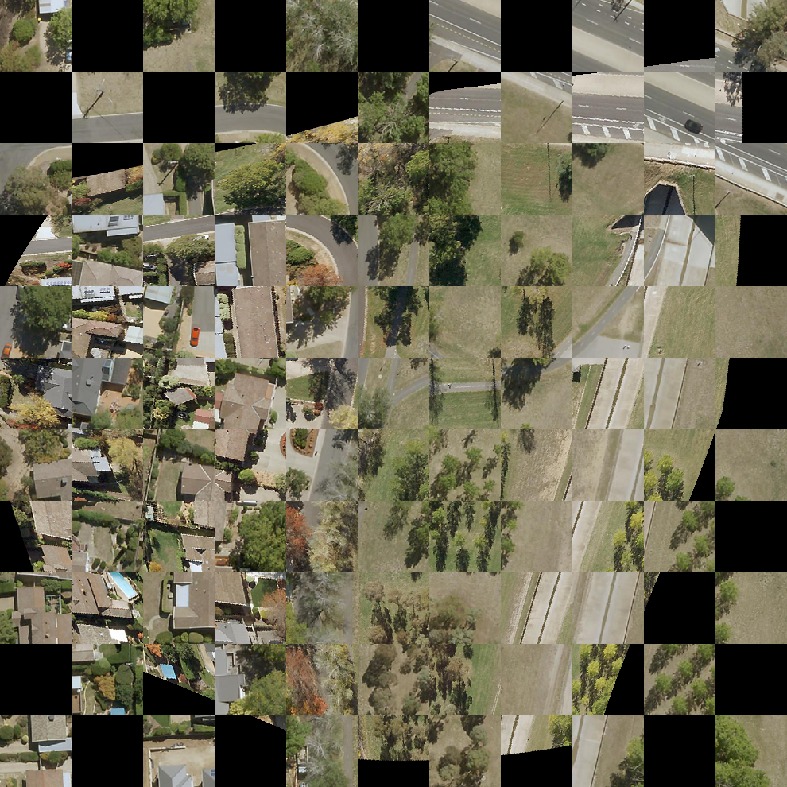} 
\end{subfigure} &
\begin{subfigure}[b]{0.24\textwidth}
\includegraphics[width = \textwidth]{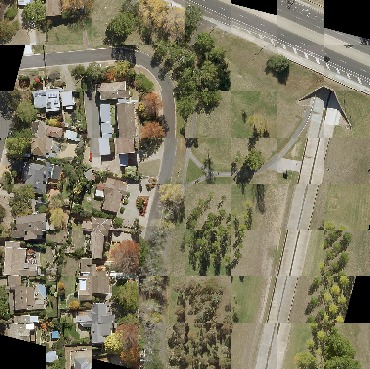} 
\end{subfigure} \\[0.3cm]

\begin{subfigure}[b]{0.24\textwidth}
\includegraphics[width = \textwidth]{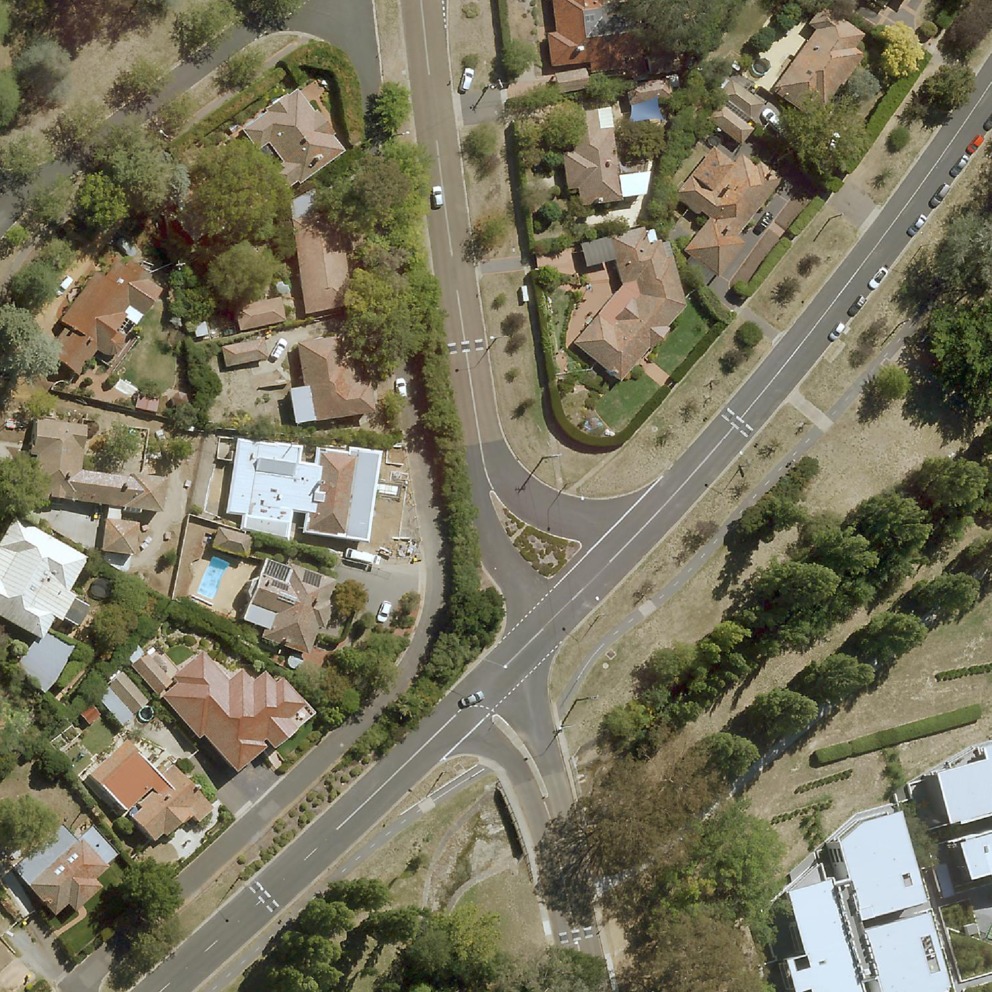} 
\end{subfigure} &
\begin{subfigure}[b]{0.24\textwidth}
\includegraphics[width = \textwidth]{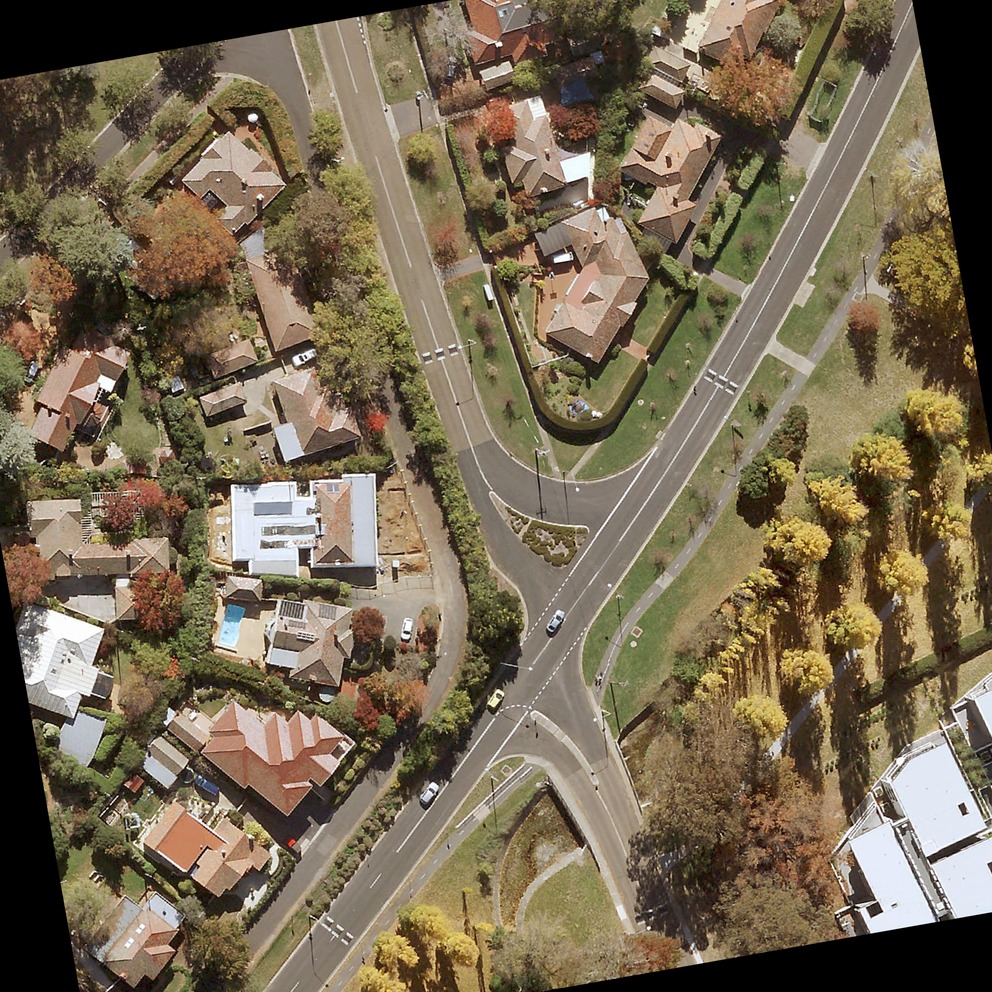} 
\end{subfigure} &
\begin{subfigure}[b]{0.24\textwidth}
\includegraphics[width = \textwidth]{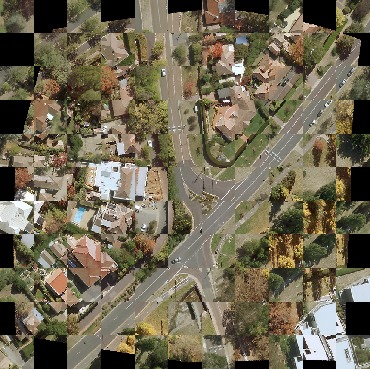} 
\end{subfigure} &
\begin{subfigure}[b]{0.24\textwidth}
\includegraphics[width = \textwidth]{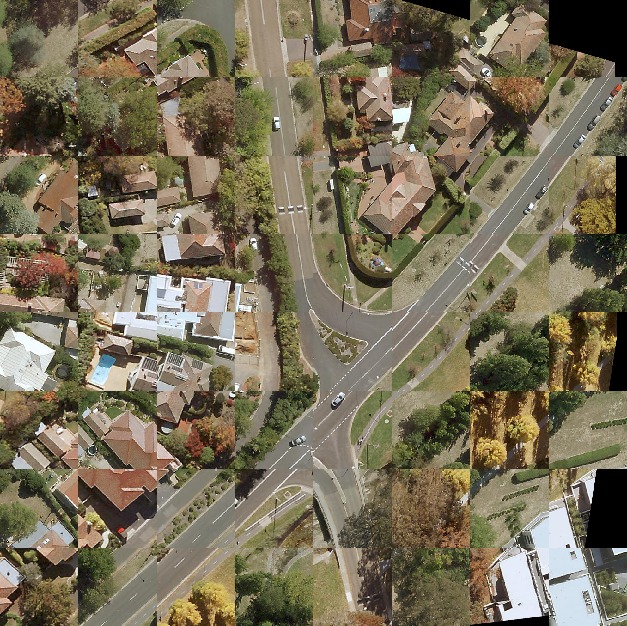} 
\end{subfigure} \\[0.3cm] \hline

\begin{subfigure}[b]{0.24\textwidth}
\includegraphics[width = \textwidth]{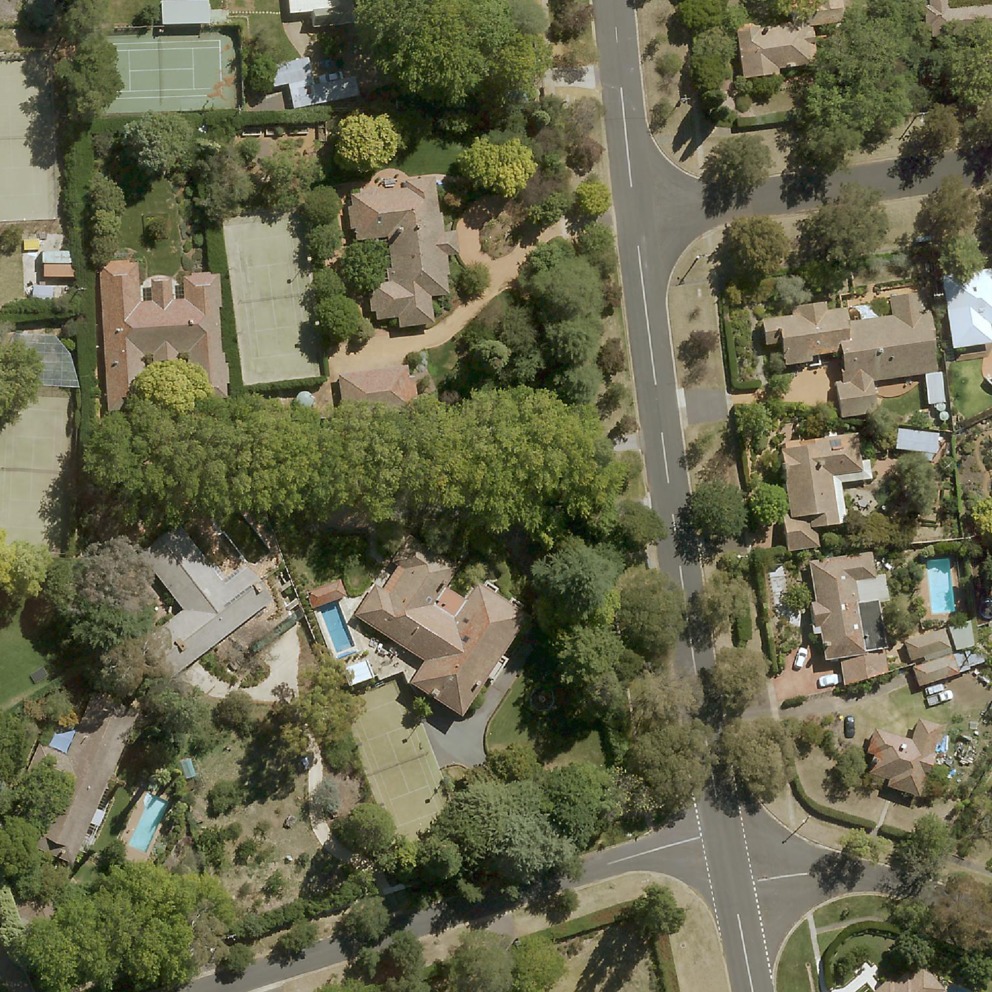} 
\end{subfigure} &
\begin{subfigure}[b]{0.24\textwidth}
\includegraphics[width = \textwidth]{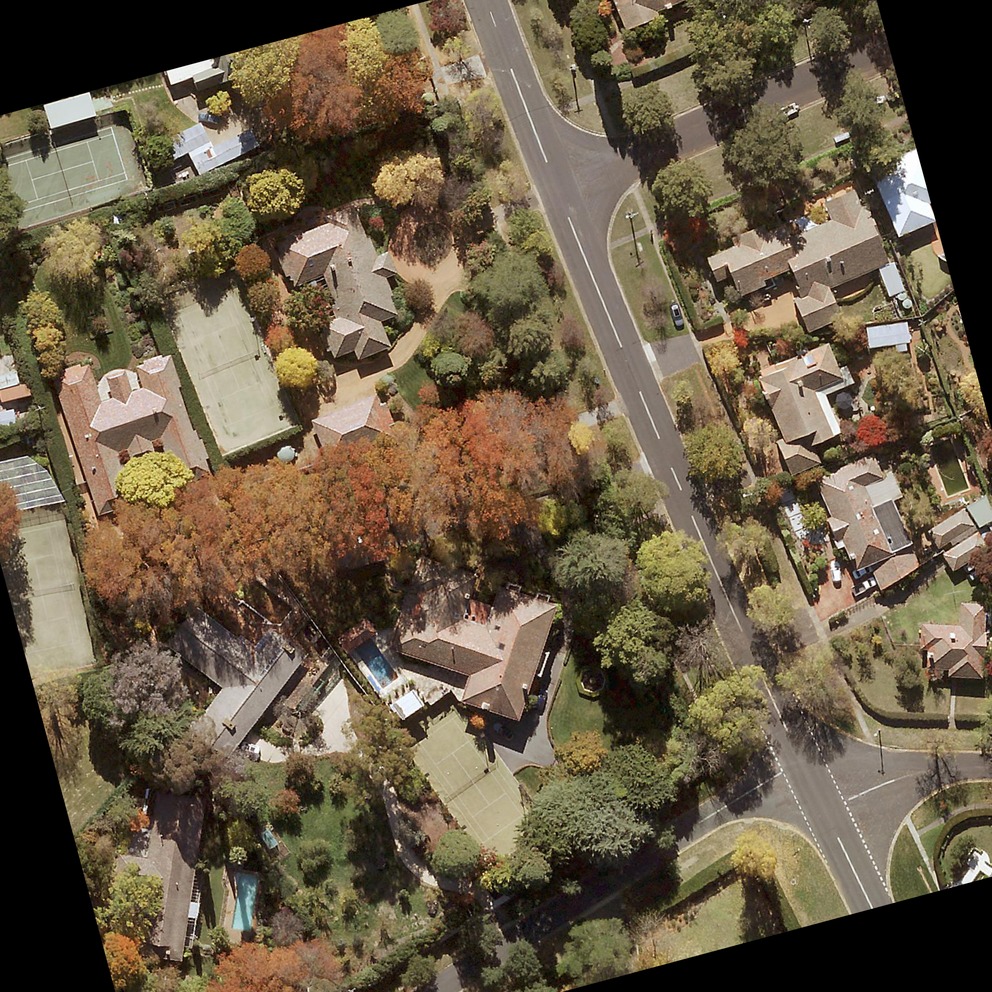} 
\end{subfigure} &
\begin{subfigure}[b]{0.24\textwidth}
\includegraphics[width = \textwidth]{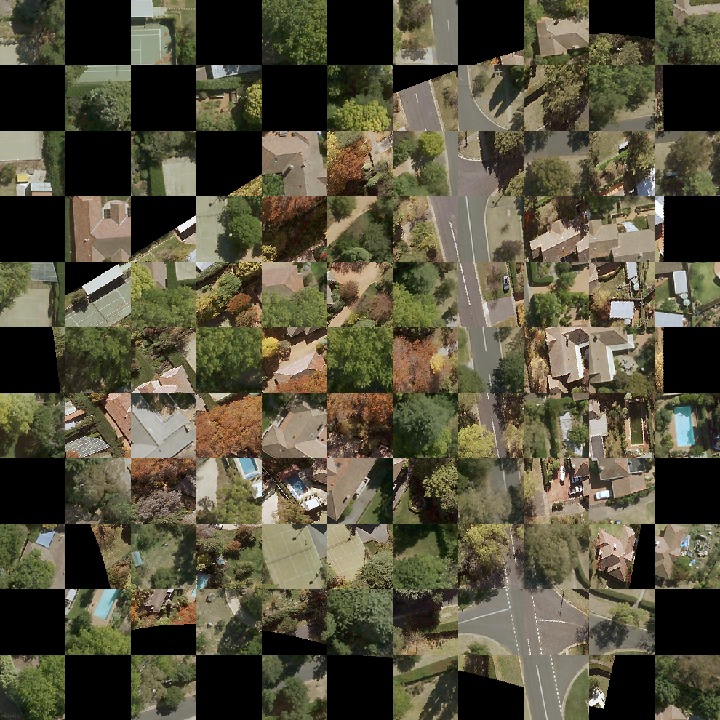} 
\end{subfigure} &
\begin{subfigure}[b]{0.24\textwidth}
\includegraphics[width = \textwidth]{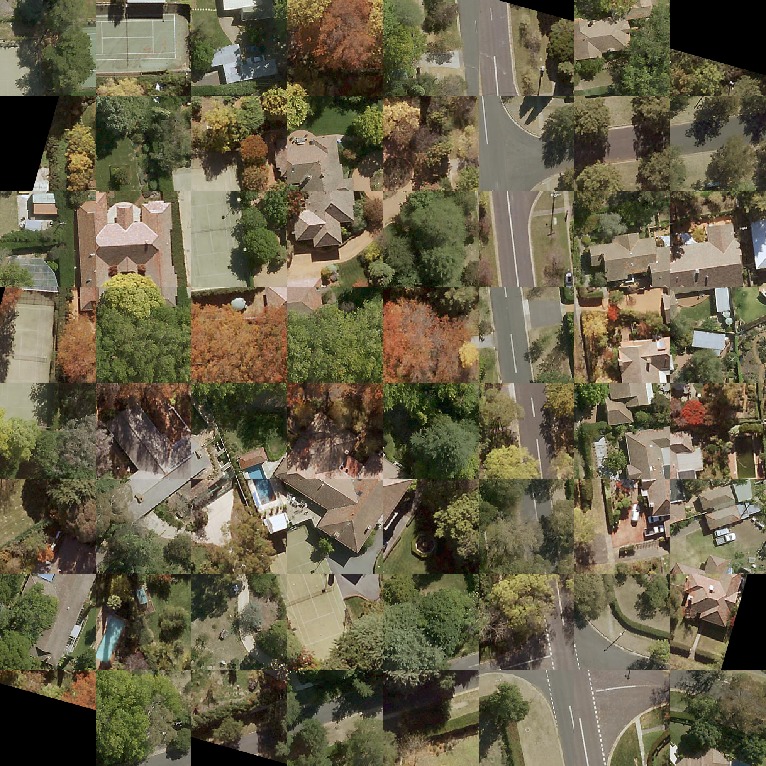} 
\end{subfigure} \\[0.3cm] \hline

\end{tabular}
\caption{Checkerboard results of registering multitemporal images using CNN-Reg and SegSF}
\label{fig:cb}
\end{figure}

As can be seen from the results in Table \ref{tab:reg_results}, SegSF performs much better than CNN-Reg for all the transformation parameters. The p-values for the tests show that the results are extremely significant. 

We believe that the performance improvement can be explained by two factors. Firstly, since CNN-Reg is based on ImageNet, it extracts universal patterns for feature matching. It does not train specifically on any aerial images but finetuning on aerial datasets, as in the case of SegSF, can provide more relevant features. However, their method cannot be finetuned with our dataset since it trains for a classification-based loss and our dataset does not contain any classes. Secondly, the addition of the semantic labels makes SegSF less sensitive to image variations. Note that both methods struggle when the rotation is increased to 20\degree and beyond.

Checkerboards of some of the images registered using these two methods are shown in Fig. \ref{fig:cb}. The query images are from 2017 and the reference images are from 2018 and the images have varied foliage. The quality of the registration can be estimated by how well the roads and buildings align on the checkerboard image. As can be seen from the figure, SegSF is able to achieve good registration results even with foliage difference.


\begin{figure}
    \centering
    \includegraphics[width=0.9\textwidth]{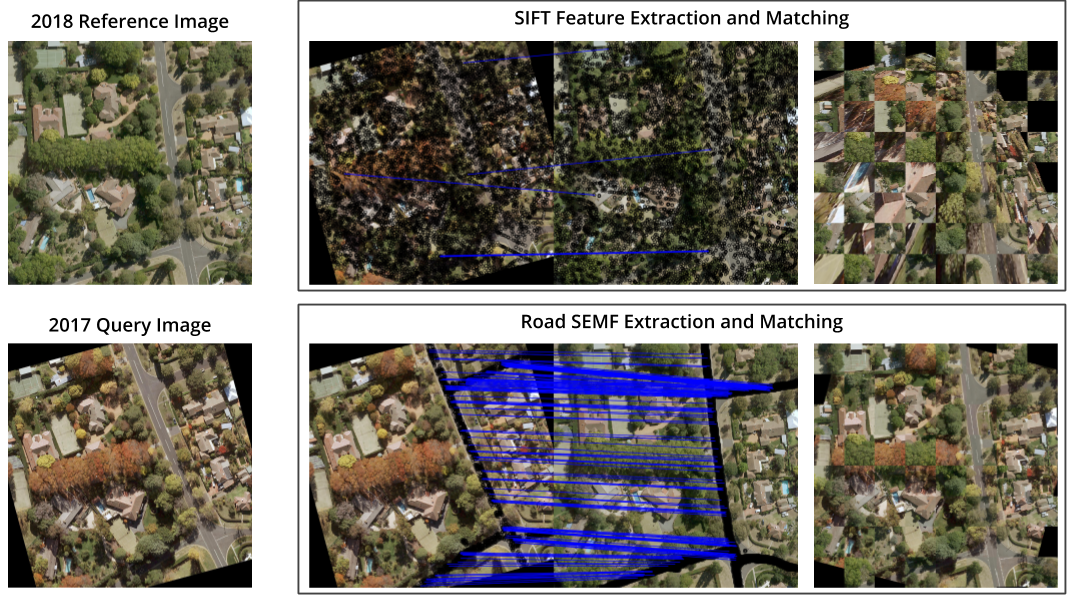}
    \caption{Matching comparison between SegSF and SIFT features. \textit{Left}: Reference image from 2018 and query image from 2017 rotated by 15 degrees with colour and shadow variations due to different times of day and year. \textit{Middle}: All keypoints extracted by the two methods are shown as black circles with the correspondences estimated by RANSAC shown with blue lines with the query image on the left and the reference image on the right. \textit{Right}: Checkerboard of the registered images.}
    \label{fig:sift_semf}
\end{figure}{}


We have also compared SegSF with SIFT features, as shown in Fig. \ref{fig:sift_semf}. As can be seen, SIFT features do not deal well with seasonal variations. Conversely, SegSF features match correctly even with large seasonal variations. Only the road features have been used in this case for the SegSF to demonstrate the relevance of semantically meaningful features. 

\section{Conclusions}

We have proposed a semantic feature extraction method for multitemporal aerial image registration to deal with variations such as changing tree foliage caused due to changing seasons, changing shadows due to different times of day, and pixel variations caused by different sensors. The features are extracted from a CNN trained for segmentation and are conditioned on the output class. We provide both quantitative and qualitative results from experiments and draw comparisons to previous work. 

Our results show that the proposed features achieve better localisation precision due to the fine resolution allowed by the use of a segmentation network as compared to a classification CNN. The features are also capable of handling temporal variations that classical features, such as SIFT, struggle with.

This work is limited in that it is only applicable to images with visible roads. However, this can be improved by increasing the number of classes for feature extraction. Further improvements in the registration accuracy can be achieved by extracting features from later layers in the network which have smaller receptive fields. Multi-scale features can also be used for dealing with scale differences.

%
%
\bibliographystyle{splncs04}
\bibliography{main}
\end{document}